\documentclass[conference]{IEEEtran}
\IEEEoverridecommandlockouts
\usepackage[utf8]{inputenc}
\usepackage{times,enumerate} 
\usepackage[usenames,dvipsnames,svgnames,table]{xcolor}
\usepackage{subcaption}
\usepackage{graphicx}
\usepackage[noadjust]{cite}

\usepackage{rotating}
\usepackage{csquotes}
\usepackage{amsmath}
\usepackage{bm}
\usepackage{tabularx}
\usepackage{stfloats}
\usepackage{threeparttable}
\usepackage{url}
\usepackage{soul}
\usepackage{threeparttable}
\soulregister\cite7
\soulregister\ref7
\soulregister\pageref7
\usepackage{amssymb}
\usepackage{soul}
\usepackage{footnote}
\usepackage{colortbl}
\usepackage{soul}
\usepackage{multirow}
\usepackage{pifont}
\usepackage{algorithmic}
\usepackage{booktabs,xcolor,colortbl}
\usepackage{color}
\usepackage{alltt}
\usepackage{hyperref}
\usepackage{enumerate}
\usepackage{siunitx}
\usepackage{epstopdf}
\usepackage{bbding}
\usepackage{pbox}
 \usepackage{amsmath}  
 \usepackage{amssymb}   
 \usepackage{amsthm}
 \usepackage[dvipsnames]{xcolor}
 
\usepackage{algorithm}

\def\BibTeX{{\rm B\kern-.05em{\sc i\kern-.025em b}\kern-.08em
    T\kern-.1667em\lower.7ex\hbox{E}\kern-.125emX}}
\begin{document}

\title{Histopathological Image Classification and Vulnerability Analysis using Federated Learning \\
%{\footnotesize \textsuperscript{*}Note: Sub-titles are not captured in Xplore and
%should not be used}
%\thanks{Identify applicable funding agency here. If none, delete this.}
}

\author{\IEEEauthorblockN{Sankalp Vyas}
\IEEEauthorblockA{\textit{Computing and Information Science} \\
\textit{Anglia Ruskin University, Cambridge}\\
sankalp.vyas@student.aru.ac.uk}
\and
\IEEEauthorblockN{Amar Nath Patra}
\IEEEauthorblockA{\textit{Radford University} \\
%\textit{Radford University}\\
Radford, USA \\
apatra@radford.edu}
\and
\IEEEauthorblockN{Raj Mani Shukla}
\IEEEauthorblockA{\textit{Computing and Information Science} \\
\textit{Anglia Ruskin University, Cambridge}\\
raj.shukla@aru.ac.uk}
}

\maketitle

\begin{abstract}
Healthcare is one of the foremost applications of machine learning (ML). Traditionally, ML models are trained by central servers, which aggregate data from various distributed devices to forecast the results for newly generated data. This is a major concern as models can access sensitive user information, which raises privacy concerns. A federated learning (FL) approach can help address this issue: A global model sends its copy to all clients who train these copies, and the clients send the updates (weights) back to it. Over time, the global model improves and becomes more accurate. Data privacy is protected during training, as it is conducted locally on the clients' devices. 

However, the global model is susceptible to data poisoning. We develop a privacy-preserving FL technique for a skin cancer dataset and show that the model is prone to data poisoning attacks. Ten clients train the model, but one of them intentionally introduces flipped labels as an attack. This reduces the accuracy of the global model. As the percentage of label flipping increases, there is a noticeable decrease in accuracy. We use a stochastic gradient descent optimization algorithm to find the most optimal accuracy for the model.  Although FL can protect user privacy for healthcare diagnostics, it is also vulnerable to data poisoning, which must be addressed.

\end{abstract}


\section{Introduction}
 In machine learning (ML), models are trained that use patterns in data to make predictions and decisions. A large amount of data is collected from multiple sources and aggregated in one location for training and evaluation. ML has potential in healthcare: By training models on medical data, tailored treatment regimens can be developed, epidemic diseases can be predicted, and trends or patterns in medical data can be identified, leading to new medicines or cures \cite{alanazi2022using, qayyum2020secure}. However, collecting and merging data can put the privacy of individuals and organizations at risk, as they may contain confidential or personal details. Additionally, transmitting massive data via the Internet can be slow and costly.

Federated learning (FL) \cite{mcmahan2017communication} overcomes these challenges by training local models on devices without data exchange. Instead of actual data, the global model is sent to these decentralized data sources and trained remotely. Additionally, FL enables model training using data controlled by numerous organizations, each of which may have different privacy concerns. Therefore, organizations can still benefit from the benefits of integrated data while protecting their privacy.

%The generated model is then communicated back to a central point for additional refinement and aggregation. This approach has the potential to maintain the privacy and security of individual devices and to manage large-scale and non-IID data. The Google Gboard keyboard app is an example of federated learning in action because it trains its predictive text model on users' devices rather than Google's servers. This prevents the centralization of private user data and allows more effective model training and updates \cite{hard2018federated}.

The importance of FL is especially pronounced when it comes to sensitive data, such as healthcare records, that are available from multiple sources. For example, automated classification of skin lesions assists dermatologists in making faster and more accurate decisions based on ML \cite{mehr2022skin}. Although ML-based systems show a decent performance in automatically classifying skin-lesion images, they need a huge amount of data that needs to be gathered for training such models. Such data is available in different hospitals and at different locations. Therefore, one of the problems encountered while developing such an automated tool is that the skin-lesion data are available across different hospitals. These data may not be shared between them due to privacy concerns and regulations. 

To solve this issue, we develop a privacy-preserving Machine Learning method for the automated classification of skin lesions that employs federated learning. The advantage of the proposed approach is that the raw patient data need not be shared with the AI service developer, and hence patient privacy is preserved. The proposed approach makes it easy to develop an AI model using a large dataset that is spatially located.  Although the proposed approach is beneficial, there is a security risk of malicious or dishonest hospitals, as they could introduce vulnerabilities into the system by sending incorrect updates. One of the strategies that they can carry out is training their local model after flipping the labels to diminish the effectiveness of the global model. Therefore, we also investigate the vulnerabilities of the proposed approach for skin lesion classification that needs to be tackled for the practical realization of such a distributed learning system. 

%The main contributions of this work are (1) Skin lesion dataset usage; (2) The demonstration/result analysis of the effect of data poisoning on the model accuracy; and (3) ...

The rest of the paper is organized as follows. Section 2 describes previous research based on FL and label-flipping attacks. The details of the preprocessing, preparation of the dataset, and the setup for building the FL model will be presented in Section 3. Furthermore, this section provides a description of the technologies used in this research. Section 4 discusses the implementation of the FL model and demonstrates the label-flipping attack. Section 5 compares the outcomes with the earlier studies in this similar area. In Section 6 the future work for this research is discussed and Section 7 concludes the work.

\section{Literature Review}
 This section examines the motivations and challenges of privacy-preserving ML, the current state of privacy-preserving machine learning, focusing on FL and distributed machine learning, and state-of-the-art skin lesion classification algorithms. 

\subsection{Motivations and Challenges for Privacy Protection in ML}
There are many reasons why privacy is important in ML, but ethical concerns are a significant factor. The use of personal data in ML models raises concerns about possible misuse of sensitive information, including financial or medical records \cite{dwork2014algorithmic}. This can result in adverse outcomes for individuals, such as discrimination and loss of autonomy \cite{barocas2016big}. 
Maintaining privacy in ML is crucial due to legal obligations. The European Union (EU), for instance, enforces strict regulations, including the General Data Protection Regulation (GDPR), which restrict the usage of personal data \cite{eu}. The purpose of these regulations is to safeguard the privacy of individuals and grant them authority over their personal information. Non-compliance with these guidelines may lead to significant penalties and damage to a business's reputation. 
Despite these motivations, protecting privacy in ML presents numerous obstacles. One difficulty is the compromise between privacy and utility. Effective ML models require enormous volumes of data, yet the collection and use of these data can compromise individuals’ privacy \cite{shokri2017membership}. Maintaining privacy while meeting the demand for data can be a difficult task to balance. Another challenge is to ensure privacy protection against attacks, including membership-inference attacks. These attacks aim to determine whether a person's data was used in training a model \cite{hu2022membership}.
The complexity of privacy-preserving procedures is another obstacle. It can be difficult to modify ML algorithms that operate on raw data without compromising their efficiency and privacy protection \cite{chaudhuri2011differentially}. Implementing new techniques can be a time-consuming and resource-intensive process. 

\subsection{Privacy-Preserving ML} 
 Privacy-preserving ML pertains to the creation of algorithms and protocols that allow the analysis of data while upholding individuals' privacy. The use of privacy-enhancing technologies (PETs) that protect privacy through technical means, such as encryption and anonymization, is one way to achieve privacy protection in ML \cite{soykan2022survey}. On the other hand, differential privacy achieves privacy preservation by adding noise to the data to obscure the contribution of individual data points \cite{dwork2006calibrating}. Another technique uses secure multiparty computation (SMC), which allows for the computation of functions over distributed data without the need to share the raw data \cite{yao1982protocols}. FL is a decentralized ML method that enables model training using distributed data without the need to centralize or share raw data \cite{mcmahan2017communication}. 
%\subsection{FL}
%FL is a decentralized ML approach that allows models to be trained on decentralized data without the need to centralize or distribute the raw data \cite{mcmahan2017communication}. In FL, a central server sends the model updates to the participating devices, which then train their respective local models using their local data and return the updated model parameters. The central server then compiles the model changes and distributes the new model to all participating devices. This technique is repeated until the model is converged. 
%FL has various privacy-preserving benefits. First, it enables models to be trained on decentralized data, eliminating the need to exchange or centralize raw data. This prevents the centralization of the gathering and processing of sensitive personal data, which could endanger privacy. Second, FL enables the training of models using data that may be subject to different restrictions, as the data remain on the local devices and are not transmitted to the central server. 

\subsection{Skin Lesion Image Classification}
ML has been used to detect and classify skin lesions in a number of studies. Gessert et al. propose a deep learning model, consisting of a convolution neural network (CNN) and a dense neural network, for classification based on dermoscopic images and additional patient metadata \cite{gessert2020skin}. The classification method designed by Yao et al. \cite{yao2021single} consists of modified deep convolution neural networks (DCNN) and a novel Multi-weighted New Loss function that counters class imbalance, improves the accuracy of key classes, and reduces outlier interference. 
Zhang et al. propose an attention residual learning convolutional neural network (ARL-CNN) model that combines residual learning and a novel attention learning mechanism to improve the discriminative representation ability of DCNNs \cite{zhang2019attention}. Wang et al. present a lightweight two-stage framework, called G-DMN \cite{wang2023classification}, that uses CycleGAN for data augmentation and Dense-MobileNetV2 (DMN) for classification. Mahbod et al. develop a fully automatic computerized classification method \cite {mahbod2019skin} that fuses deep features from several robust CNNs, including AlexNet, VGG16, and ResNet-18, and different abstraction levels. 

The aforementioned research employs a centralized approach in which skin lesions are collected from different locations, compromising user privacy. In contrast,  we propose a FL-based approach to preserve user privacy that does not collect raw data in one place. Instead, the intelligence in the form of trained models is shared between and combined to develop the ML model.  Additionally, we analyze the vulnerability of such a system due to the presence of possible malicious clients.

\section{Methodology}
This section describes the proposed FL-based approach along with the system model and the data poisoning vulnerability scenario.

\subsection{System model}
\begin{figure*}
    \centering
    \includegraphics[width=0.75\linewidth]{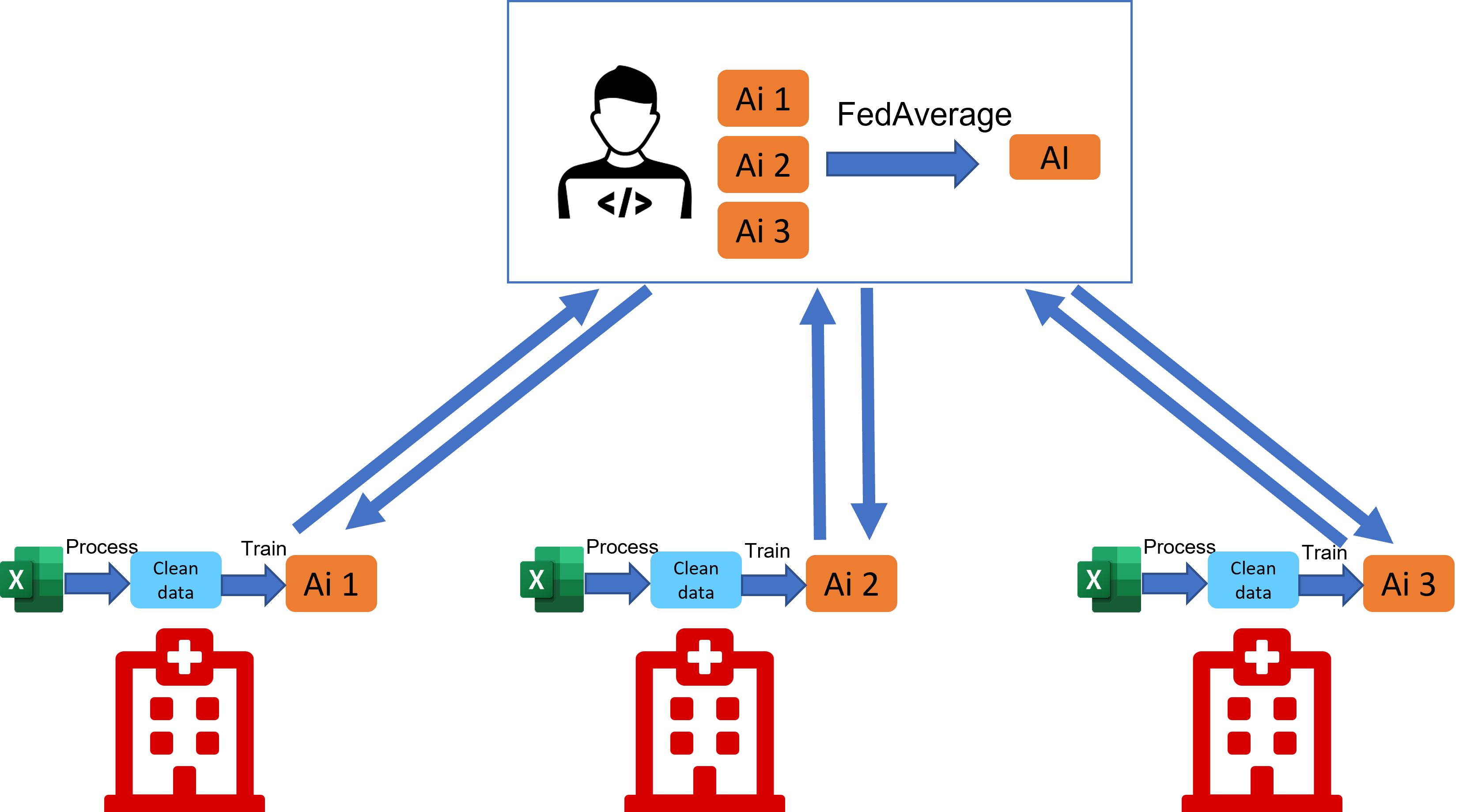}
    \caption{The distributed learning environment for the proposed Histopathological image classification}
    \label{fig:model}
\end{figure*}

Fig.~\ref{fig:model} presents the system model of the proposed distributed learning environment for histopathological image classification. A hospital $h[i]$ holds raw private images of its patients, which it processes for clean images. Based on the clean dataset, $h[i]$ can train its local model $A_{i}$. The application developer, who may represent a company providing Software as a Service (SaaS), is at the other end and does not have access to the raw data. However, the local models from different hospitals are shared through a secure channel with the application developer aggregator. The developer combines these models to create a global model $AI$ that is deployed in the cloud or on the edge as a SaaS platform. This platform can be used by any hospital or individual, regardless of whether they participate in model training. The business model may allow entities participating in the training process to use the platform for free or at a discounted price, while entities not participating in the training process must pay full subscription rates. With more entities participating in the training process and sharing intelligence through their respective trained models, a better global model is developed. 

As new data are continuously collected, participants in the training process can retrain the model, using the global model's weights as initial weights. Over time, the SaaS platform can improve itself as it captures disparate and varied distributions of data.

\subsection{Proposed FL-based approach}

As the proposed methodology uses FL, the training dataset $D$ is not fully available on a centralized server but on multiple participants $h[i]$, who have their own private dataset $D_i$. Each $h[i]$ executes stochastic gradient descent (SGD) locally on its own data and then sends its updated model parameters $\theta_{r,i}$ to the aggregator. The aggregator combines the parameters received to create global parameters $\theta_r$, which are broadcast to all participants $h[i]$. This process is repeated for a fixed number of rounds $R$ and the final model $AI_{M}$ is created using the global parameters of the last round $\theta_R$.

The deep neural network (DNN) model takes the features $f_i$ and outputs the predicted probabilities $p_i$ for each skin lesion class label $c_i$ in a set of all possible class values $C$. The predicted class for a given instance $x_i$ is determined by finding the class label with the highest predicted probability. The goal of training the DNN is to find the model parameters $\theta$ that minimize the chosen loss function $L$ using iterative SGD. In SGD, a batch of samples $B$ is selected from the dataset $D$, and the corresponding gradient $gB$ is computed. The model parameters $\theta$ are then updated in the direction of the negative gradient. This process is repeated for multiple epochs.

%Federated learning systems allow for the training of the global model without the sharing of private data in its raw form. Instead, only the update of the model parameters is shared by the participants. Assume a deep neural network model like a multilayer perceptron model. Each layer of an MLP gets the input from the previous layer, processes it, and forwards it to the next layer until the output is generated. In a multilayer perceptron network model, each neuron in the input and output layers is coupled to every neuron in the hidden layers. 
%The input to an MLP is passed through the input layer and the hidden layers, where it is converted and modified by the weights and biases of the connections between the neurons. The adjusted input is finally transmitted through the output layer to generate the MLP output. In summary, the inputs and the outputs in an MLP are combined through the connections between the neurons in the different layers of the network. The weights and biases of these connections determine how the input is modified as it moves through the hidden layers to generate the output.

In the process of training a DNN, dataset $D$ with a set of features $f_i$ and class labels $c_i$ is used, along with a loss function $L$. This loss function is given by

\begin{equation}
    \label{eq:loss function}
    L_{total} = \frac{1}{n} \sum^n _i L (\theta, x_i)
\end{equation}

The proposed method uses the Federated Averaging (FedAvg) model, a distributed machine learning algorithm designed for model training in a federated setting, where multiple clients or devices, each with their own training data and computational resources, train a shared model. FedAvg’s goal is to train a model capable of making accurate predictions on the data of all clients, even though the data of each client are only a subset of the total data \cite{tolpegin2020data}. The hospitals act as the clients in the proposed method. 

The overview of the FedAvg algorithm for the proposed histopathological image classification is described below.
\begin{itemize}
    \item \textit{Initialization} The algorithm begins by initializing the global model $AI$ with random weights and biases. $AI$ is then distributed to each hospital $h[i]$ which uses it as its local model $A_{i}$.
    \item \textit{Communication rounds} The algorithm goes through a series of communication rounds (federated learning rounds) with the following steps in each round
    \begin{itemize}
       \item \textit{Local training} Each $h[i]$ chooses a subset of its data $D_{i}$ (mini-batch) to train with. Then, using the selected data, it performs one or more gradient descent steps on its local model $A_{i}$.
        \item \textit{Averaging} Each $h[i]$ updates the model parameters $\theta_{r,i}$ and sends them to the aggregator. The received models are then averaged together by the aggregator to create a new global model $AI$.
    \item \textit{Update} $AI$ is then returned to the $h[i]$s to use as local models $A_{i}$s.
    \item \textit{Repeat} The process repeats until $AI$ reaches a certain level of accuracy of a stopping criterion.
    \end{itemize}
\end{itemize}

The averaging step in the FedAvg algorithm is crucial, as it updates $AI$ by taking the average of the $A_{i}$s. This allows the algorithm to account for differences in data across $h[i]$s, which can help improve $AI$'s generalization performance.

\subsection{Malicious clients}
Although the aforementioned approach preserves user privacy, there exists a vulnerability if one or more $h[i]$s provide incorrect model updates. A hospital $h[i]$ may adopt the following steps to alter the outcome of the global model $AI$

\begin{enumerate} 
    \item Select $k$ samples from the labeled dataset where $k$ is the $p$ percentage of the selected labels
    \item For each $k$ replace the skin-lesion class with an alternate random skin-lesion class
    \item Repeat step 2 for all k
    \item Train the local model $A_{i}$ on the substituted label
    \item Send the trained model to the aggregator for Fedaveraging.
\end{enumerate}

\section{Performance Evaluation and Results}
In this section, we provide the details of implementation that include the FL environment setup, DNN architecture, data collection, analysis, visualization, dataset pre-processing, and evaluation methods employed to obtain the results. We analyze the results before and after the flipping of the labels and do a comparative study.

\subsection{Implementation details}
%\raj{Done this subsection. Need to be refined.}
%\anp{Refined and added new comments. Quick question: we have not mentioned the list of \textit{contributions} and explanation of how our work is novel in the introduction (and related work) section. Will it affect our paper acceptance?}

\subsubsection{FL setup} The FL system is implemented in the TensorFlow library \cite{tensorflow2015-whitepaper}, available in Python. $N$ represents the number of participating hospitals in the FL setup which is set to 10, whereas there is one aggregator. Additionally, independent and identity-distributed data (IID) are used. 

%This means that the data for each of our clients is the same, and it is drawn from the same distribution. We did not use our testing dataset as it was kept for testing purposes. Our model ran for a total of 100 communication rounds, which is nothing but 100 training rounds. 

\subsubsection{Deep neural network architecture} In the setup, we use a multilayer-perceptron model (MLP) to classify images from the dataset into one of the seven classes. The MLP model consists of three fully connected (dense) layers with 200 units in each. Relu activation is used in the hidden layer and softmax in the output layer. 

Each hospital $h[i]$ is responsible for training its local model $A_{i}$ on its local dataset $D_{i}$. We use the SGD optimizer to train the AI model. We train the model for 100 communication rounds, during which the updates from each $h[i]$ are aggregated on the central server and used to update the global model $AI$. To measure the error of the model during training, we use the categorical\_crossentropy loss function. To evaluate the performance of the model, we use the \textit{accuracy} metric, which measures the proportion of correct predictions made by the model.

%Overall, our federated learning setup allows us to train a model on a large, distributed dataset while maintaining the privacy of the individual client data. By aggregating the updates from the clients and using the SGD optimizer with the specified parameters, we can improve the performance of the global model and achieve better results compared to training on a single, centralized dataset.

%\raj{We need to mention the following citation for HAM dataset, not Kaggle https://www.nature.com/articles/sdata2018161}

\subsubsection{Data collection} 
\begin{figure}
    \centering
    \includegraphics[width=0.95\linewidth]{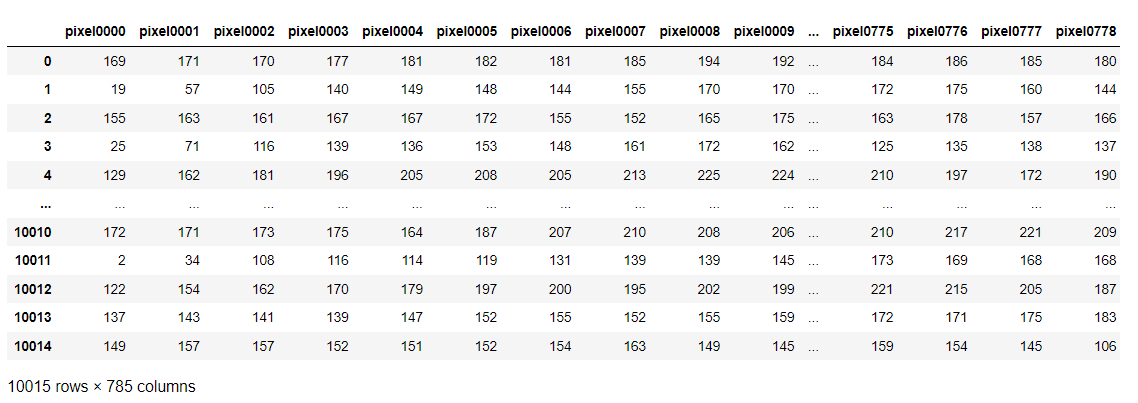}
    \caption{Sample skin lesion data}
    \label{fig:samples}
\end{figure}

We use the HAM10000 dataset that consists of a total of 10,015 samples that belong to seven different classes \cite{tschandl2018SD}. Fig. \ref{fig:samples} shows the image data for all 10,015 samples and the seven classes.

\subsubsection{Data analysis and visualization} 
\begin{figure}
    \centering
    \includegraphics[width=0.85\linewidth]{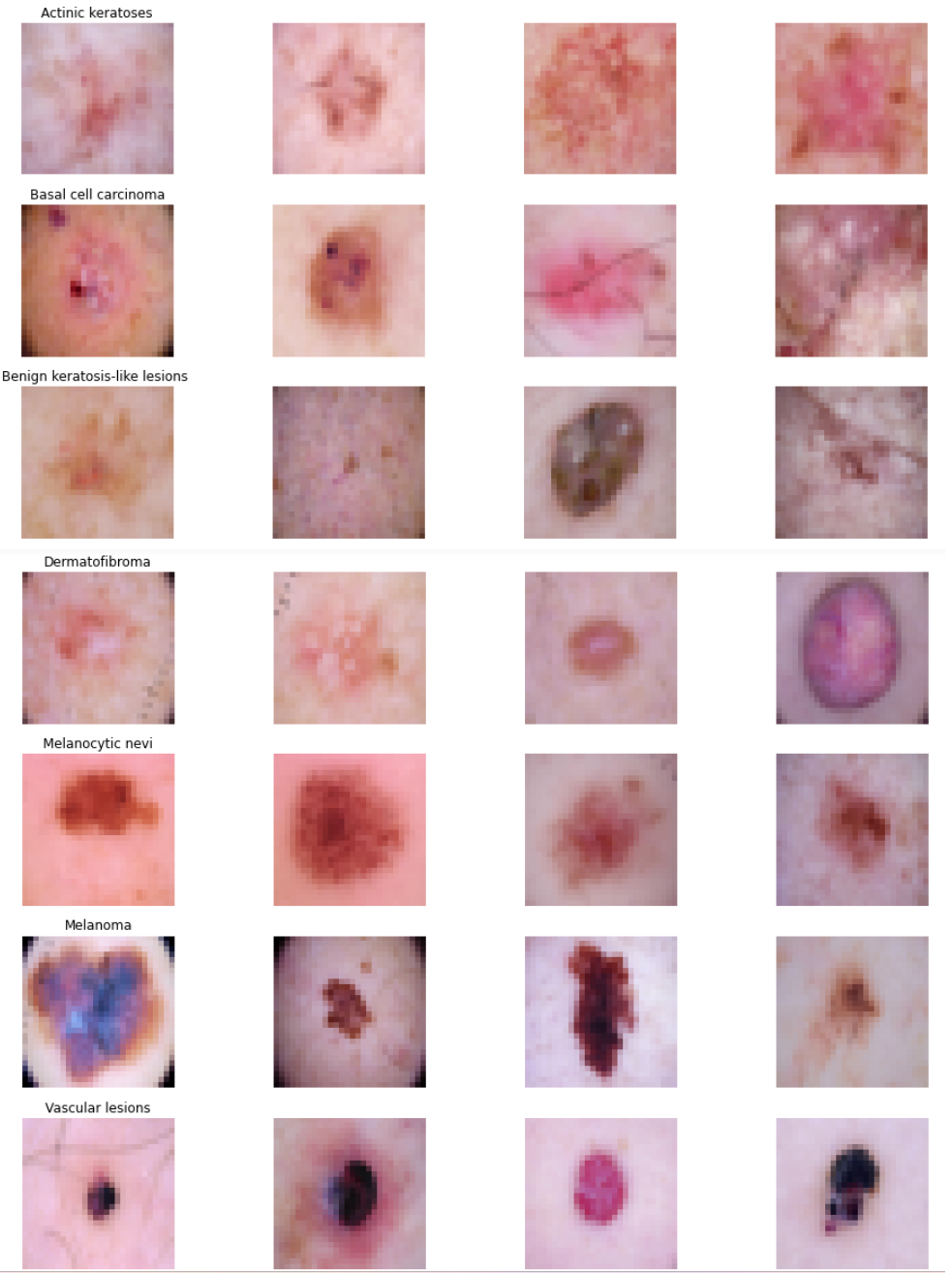}
    \caption{Types of skin lesions and their samples}
    \label{fig:lesions}
\end{figure}

The seven classes of skin cancer lesions in the HAM10000 dataset are Melanocytic nevi (nv), Melanoma (mel), Benign keratosis-like lesions (bkl), Basal cell carcinoma (bcc), Actinic keratoses (akiec),  Vascular lesions (vasc), and Dermatofibroma (df). Fig. \ref{fig:lesions} shows some of the samples of the seven different types of lesions.

%\begin{figure}
%    \centering
%    \includegraphics[width=0.9\linewidth]{pie.png}
%   \caption{Distribution of the lesions}
%    \label{fig:pie}
%\end{figure}

%\anp{Is Fig. \ref{fig:pie} referred anywhere?}

\subsubsection{Pre-processing of the dataset}
In the pre-processing stage, each pixel in the image is represented by a single greyscale value (ranging from 0 to 255). When these images are stored in a numerical array, they will have 784 elements (28x28), each representing the greyscale value of a single pixel. Thus, there are a total of 784 columns and 10,015 rows. Table  \ref{table:imageData} shows the image columns containing values between 0 and 1. The images have been converted to a 1D array as we use MLP to build our FL model. Pixel000, pixel001..., pixel0784, etc. represent the number of pixels while 0,1,2,3,4…, 10,014 indicate the total number of samples. 

%\begin{figure}
%    \centering
%    \includegraphics[width=0.9\linewidth]{imageData.png}
%    \caption{– Image data after getting scaled between 0 and 1}
%    \label{fig:imageData}
%\end{figure}

\begin{table*}
\caption{Image data after getting scaled between 0 and 1}
\label{table:imageData}
\centering
\begin{tabular}{|p{0.5in}|p{0.5in}|p{0.5in}|p{0.5in}|p{0.5in}|p{0.5in}|p{0.5in}|p{0.5in}|p{0.5in}|p{0.5in}|}
\hline
	\textbf{pixel0000} & \textbf{pixel0001} & \textbf{pixel0002} & \textbf{pixel0003} & \textbf{pixel0004} & \textbf{pixel0005} & \textbf{pixel0006} & \textbf{pixel0007} & \textbf{pixel0008} & \textbf{pixel0009}\\ \hline 

0.674510 & 0.670588 & 0.678431 & 0.686275 & 0.643137 & 0.733333 & 0.811765 & 0.823529 & 0.815686 & 0.807843\\ \hline
        0.007843 & 0.133333 & 0.423529 & 0.454902 & 0.447059 & 0.466667 & 0.513725 & 0.545098 & 0.545098 & 0.568627\\ \hline
        0.478431 & 0.603922 & 0.635294 & 0.666667 & 0.701961 & 0.772549 & 0.784314 & 0.764706 & 0.792157 & 0.780392\\ \hline  
        0.537255 & 0.560784 & 0.552941 & 0.545098 & 0.576471 & 0.596078 & 0.607843 & 0.596078 & 0.607843 & 0.623529\\ \hline
        0.584314 & 0.615686 & 0.615686 & 0.596078 & 0.607843 & 0.596078 & 0.603922 & 0.639216 & 0.584314 & 0.568627\\ \hline
\end{tabular}
\end{table*}

\subsubsection{Evaluation methods} The metrics used for evaluation are precision, recall, f1-scores, and accuracy (Table \ref{table:recall}). Precision is the number of true positive predictions divided by the number of true positive plus false positive predictions. The number of true positive predictions divided by the number of true positive plus false negative predictions is recall. Accuracy is the percentage of correct predictions made by the model out of all the predictions made by the model, while the F1 score is the harmonic mean of precision and recall.

\begin{table}
    \centering
    \caption{Evaluation Metrics}
    \label{table:recall}
    \begin{tabular}{ccc}\hline
	\textbf{Evaluation Metrics} & \textbf{Formula} \\ \hline
	Recall & $\frac{True Positive (TP)}{True Positive (TP) + False Negative (FN)}$ \\ \hline
        Precision & $\frac{True Positive (TP)}{True Positive (TP) + False Positive (FP)}$ \\ \hline
        F1-score & $2 \times \frac{Precision \times Recall}{Precision + Recall}$ \\ \hline
        Accuracy & $\frac{True Positive (TP) + True Negative (TN)}{Total number of predictions}$ \\
        \hline
    \end{tabular}
\end{table}

\subsection{Results}

We discuss the results in two parts, before and after the flipping of the labels.

\subsubsection{Results before flipping of the labels} The parameters taken into consideration for building the FL model are shown in Table \ref{table:parameters} while Fig. \ref{table:eval} shows the evaluation metrics for the model.

\begin{table}
    \centering
    \caption{Variation of k with p}
    \label{table:parameters}
    \begin{tabular}{cc}\hline
	\textbf{Parameters} & \textbf{Value} \\ \hline
	learning\_rate & 0.01 \\ 
        comms\_round & 100 \\
        Input shape & (784,) \\
        Number of classes & 7 \\
        Bs (batch size) & 32 \\
        num\_clients & 10 \\
        loss & categorical\_crossentropy \\
        metrics & accuracy \\
        momentum & 0.9 \\
        \hline
    \end{tabular}
\end{table}

%\begin{figure}
 %   \centering
  %  \includegraphics[width=0.9\linewidth]{eval.png}
  %  \caption{Results along with classification report for the global model before the flipping of the labels}
   % \label{fig:eval}
%\end{figure}

\begin{table}
\caption{Classification report of the global model in FL setup before the flipping of the labels}
\label{table:eval}
\centering
\begin{tabular}{|p{0.65in}|p{0.4in}|p{0.3in}|p{0.4in}|p{0.35in}|}
\hline
& \textbf{precision} & \textbf{recall} & \textbf{f1-score} & \textbf{support} \\ \hline 

0 & 0.00 & 0.00 & 0.00 & 61 \\ \hline
1 & 0.33 & 0.01 & 0.02 & 96 \\ \hline
2 & 0.30 & 0.08 & 0.13 & 228 \\ \hline  
3 & 0.00 & 0.00 & 0.00 & 37 \\ \hline
4 & 0.69 & 0.99 & 0.81 & 1327 \\ \hline
5 & 0.00 & 0.00 & 0.00 & 32 \\ \hline
6 & 0.57 & 0.07 & 0.13 & 222 \\ \hline
macro avg & 0.27 & 0.16 & 0.16 & 2003 \\ \hline
weighted avg & 0.57 & 0.67 & 0.57 & 2003 \\ \hline
\multicolumn{3}{|c|}{accuracy} & 0.67 & 2003 \\ \hline

\end{tabular}
\end{table}

Fig. \ref{table:eval} shows the classification report for a multiclass classification problem, where the model is trying to predict one of the 7 different classes (0-6). The report provides precision, recall, f1 score, and support for each class.
High precision implies that the model mostly predicts the correct class while a high recall means that the model can identify most of the positive instances. As the f1-score is a weighted harmonic mean of precision and recall, it balances precision and recall and is thus particularly useful when the classes are imbalanced.
Support is the number of instances that belong to each class. The overall accuracy of the model is 67.099\%. However, the accuracy can be misleading, especially when the classes are imbalanced. It's important to look at the performance of each class separately. The model correctly identifies class 4 with a high precision of 0.69 and a recall of 0.99, which is quite good. The f1-score is 0.81 for class 4. However, the model is not performing well on the $5^{th}$ as well as the $0^{th}$ class as the precision, recall, and f1-score are all zero. This is due to the small number of instances in those classes, which makes it difficult for the model to learn to predict them correctly.
This same model was also tested on a SGD optimizer without using FL. To ensure a level playing ground, each of the hyperparameters was retained during the FL training, with the exception of the batch size. Instead of 32, the batch size in this instance was set to 320. With this configuration, the SGD will see the same number of training samples in each iteration as the global model $AI$ did for each communication round in FL with this setup. Fig. \ref{table:classification} shows the detailed report for the centralized setup.

%\begin{figure}
 %   \centering
 %   \includegraphics[width=0.9\linewidth]{classification.png}
 %   \caption{Classification report for the SGD model}
 %   \label{fig:classification}
%\end{figure}

\begin{table}
\caption{Classification report for the centralized configuration}
\label{table:classification}
\centering
\begin{tabular}{|p{0.65in}|p{0.4in}|p{0.3in}|p{0.4in}|p{0.35in}|}
\hline
& \textbf{precision} & \textbf{recall} & \textbf{f1-score} & \textbf{support} \\ \hline 

0 & 0.00 & 0.00 & 0.00 & 61 \\ \hline
1 & 0.20 & 0.01 & 0.02 & 96 \\ \hline
2 & 0.32 & 0.06 & 0.10 & 228 \\ \hline  
3 & 0.00 & 0.00 & 0.00 & 37 \\ \hline
4 & 0.68 & 0.99 & 0.81 & 1327 \\ \hline
5 & 0.00 & 0.00 & 0.00 & 32 \\ \hline
6 & 0.53 & 0.08 & 0.14 & 222 \\ \hline
macro avg & 0.25 & 0.16 & 0.15 & 2003 \\ \hline
weighted avg & 0.56 & 0.67 & 0.56 & 2003 \\ \hline
\multicolumn{3}{|c|}{accuracy} & 0.67 & 2003 \\ \hline

\end{tabular}
\end{table}

The accuracy for the centralized ML is 67.099\% while the global loss is 1.55 (Fig. \ref{table:eval}. In this configuration, the $4^{th}$ class has a recall of 0.99 and a f1-score of 0.81. The precision is 0.68. The model does not perform well on the $0^{th}$ class as well as the $5^{th}$ class due to a smaller number of samples for both of these classes.

\subsubsection{Results after flipping of labels} The same model was run again after a certain percentage of labels had been randomly altered. 2\% of the labels were changed in the beginning, and this process continued until 20\% of the labels were changed for the malicious client. The accuracy, precision, recall, and f1 scores were recorded after changing the labels from 2\% to 20\%. Table \ref{table:afterFlipping} shows that the classification report changes when a certain percentage of labels are flipped. For example, when 14\% labels are changed, there is a decrease in accuracy from 67. 099\% to 66. 101\%.

%\begin{figure}
 %   \centering
 %   \includegraphics[width=0.9\linewidth]{afterFlipping.png}
%    \caption{Results of the model after flipping 14\% of labels}
 %   \label{fig:afterFlipping}
%\end{figure}

\begin{table}
\caption{Classification report of the global model in FL setup after flipping 14\% of labels}
\label{table:afterFlipping}
\centering
\begin{tabular}{|p{0.65in}|p{0.4in}|p{0.3in}|p{0.4in}|p{0.35in}|}
\hline
& \textbf{precision} & \textbf{recall} & \textbf{f1-score} & \textbf{support} \\ \hline 

0 & 0.00 & 0.00 & 0.00 & 61 \\ \hline
1 & 0.00 & 0.00 & 0.00 & 96 \\ \hline
2 & 0.00 & 0.00 & 0.00 & 228 \\ \hline  
3 & 0.00 & 0.00 & 0.00 & 37 \\ \hline
4 & 0.66 & 0.99 & 0.80 & 1327 \\ \hline
5 & 0.00 & 0.00 & 0.00 & 32 \\ \hline
6 & 0.75 & 0.03 & 0.05 & 222 \\ \hline
macro avg & 0.20 & 0.15 & 0.12 & 2003 \\ \hline
weighted avg & 0.52 & 0.66 & 0.53 & 2003 \\ \hline
\multicolumn{3}{|c|}{accuracy}  & 0.66 & 2003 \\ \hline
\end{tabular}
\end{table}

The $0^{th}$,$1^{st}$,$2^{nd}$,$3^{rd}$, and $5^{th}$ classes had a precision, recall, and f1-score of 0.00 while the $4^{th}$ class had a precision of 0.66, a recall of 0.99 and a f1-score of 0.80. The $6^{th}$ class had a precision of 0.75, a recall of 0.03, and a f1-score of 0.05. Table \ref{table:accuracyChange} shows the changes in accuracy before the flipping of the labels and after the flipping of the labels. The table reveals that for most cases there is a drop in accuracy when a malicious client randomly flips the label.

\begin{table}
    \centering
    \caption{Change in accuracy}
    \label{table:accuracyChange}
    \begin{tabular}{|p{0.5in}|p{0.5in}|p{0.5in}|p{0.5in}|p{0.5in}|}\hline
	\textbf{Percentage of labels flipped} & \textbf{Global accuracy} & \textbf{Global accuracy after label flipping} & \textbf{Centralized setup accuracy} & \textbf{Centralized setup accuracy after label flipping}\\ \hline
	2\% & 67.099 & 67.049 & 67.499 & 67.399  \\ 
        4\% & 67.099 & 67.099 & 67.499 & 67.149 \\
        6\% & 67.099 & 66.600 & 67.499 & 67.049 \\
        8\% & 67.099 & 66.600 & 67.499 & 67.000  \\
        10\% & 67.099 & 66.350 & 67.499 &66.800  \\
        12\% & 67.099 & 66.151 & 67.499 &66.400  \\
        14\% & 67.099 & 66.101 & 67.499 &66.400  \\
        16\% & 67.099 & 66.051 & 67.499 &66.301  \\
        18\% & 67.099 & 66.900 & 67.499 &67.149 \\
        20\% & 67.099 & 67.249 & 67.499 &67.299 \\
        \hline
    \end{tabular}
\end{table}

%\textbf{Resources used} Google Colab was used for implementing the experiments with the settings shown in Table \ref{table:colab}.

%\begin{table}
    %\centering
    %\caption{Google Colab settings}
    %\label{table:colab}
    %\begin{tabular}{cc}\hline
	%\textbf{Hardware type} & \textbf{Configuration} \\ \hline
	%Hard disk & 1TB \\ 
        %GPU & Tesla T4 \\
        %CPU & Intel i5 8th Gen \\
        %RAM & 83.48 GB \\
        %\hline
    %\end{tabular}
%\end{table}

%Pandas 1.3.5 and NumPy 1.21.6 libraries were used exclusively to manipulate data. Scikit-learn 1.0.2 was utilized to calculate the evaluation metrics and pre-processing the data. TensorFlow 2.9.2 library was used to build local and global models. %Table \ref{table:Software} provides the version numbers for the libraries used.

%\begin{table}
%    \centering
%    \caption{Software Versions}
%   \label{table:Software}
%    \begin{tabular}{cc}\hline
%	\textbf{Software} & \textbf{Version} \\ \hline
%	Numpy & 1.21.6 \\ 
 %       Scikit-learn & 1.0.2 \\
%        TensorFlow & 2.9.2 \\
 %       Python & 3.8.16 \\
%        Pandas & 1.3.5 \\
%        \hline
%    \end{tabular}
%\end{table}

%\anp{What content should be used from the Appendix? I did not add/copy anything from it.}

\section{Conclusion}
In this paper, we built a privacy-preserving federated learning model and showed the vulnerability of this model by attacking it through a malicious client. We demonstrated that this malicious client changed a certain number of samples which led to a decrease in the accuracy of the global model. We observed that for most of the cases, there is a decrease in the performance as well as the accuracy of the model. We also compared the results to a Stochastic gradient descent optimizer in a centralized learning setup. Moreover, we also compared the results before and after changing the labels and found a decrease in the performance and the accuracy of the model. In the future, rather than randomly flipping the label, optimization methods to maximally affect the accuracy could be analyzed.

\bibliographystyle{ieeetr}
\bibliography{references}
\end{document}